\documentclass[10pt, a4paper]{article}

\PassOptionsToPackage{table}{xcolor}

\usepackage[final]{lrec2026} 
\definecolor{top1}{RGB}{166,166,255}
\definecolor{top2}{RGB}{204,204,255}
\definecolor{top3}{RGB}{242,242,255}
\usepackage{multirow}
\usepackage[utf8]{inputenc}
\usepackage{kotex}
\usepackage{subcaption} 

\usepackage{colortbl}
\definecolor{errorbg}{RGB}{255,240,240}   
\definecolor{goldbg}{RGB}{235,245,255}    
\definecolor{errorfg}{RGB}{180,40,40}
\definecolor{goldfg}{RGB}{30,90,160}

\title{BankMathBench: A Benchmark for Numerical Reasoning \\ in Banking Scenarios}

\name{Yunseung Lee$^{*12}$\thanks{* These authors contributed equally to this work.}, Subin Kim$^{*1}$, Youngjun Kwak$^{1}$, Jaegul Choo$^{2}$}

\address{$^{1}$KakaoBank Corp., South Korea \\
        $^{2}$Korea Advanced Institute of Science and Technology, South Korea \\
         \{yun.lee, luna.ns, vivaan.yjkwak\}@lab.kakaobank.com, jchoo@kaist.ac.kr}

\abstract{
Large language models (LLMs)-based chatbots are increasingly being adopted in the financial domain, particularly in digital banking, to handle customer inquiries about products such as deposits, savings, and loans. However, these models still exhibit low accuracy in core banking computations—including total payout estimation, comparison of products with varying interest rates, and interest calculation under early repayment conditions. 
Such tasks require multi-step numerical reasoning and contextual understanding of banking products, yet existing LLMs often make systematic errors—misinterpreting product types, applying conditions incorrectly, or failing basic calculations involving exponents and geometric progressions. However, such errors have rarely been captured by existing benchmarks.
Mathematical datasets focus on fundamental math problems, whereas financial benchmarks primarily target financial documents, leaving everyday banking scenarios underexplored. To address this limitation, we propose BankMathBench, a domain-specific dataset that reflects realistic banking tasks. BankMathBench is organized in three levels of difficulty—basic, intermediate, and advanced—corresponding to single-product reasoning, multi-product comparison, and multi-condition scenarios, respectively. When trained on BankMathBench, open-source LLMs exhibited notable improvements in both formula generation and numerical reasoning accuracy, demonstrating the dataset’s effectiveness in enhancing domain-specific reasoning. With tool-augmented fine-tuning, the models achieved average accuracy increases of 57.6\%p (basic), 75.1\%p (intermediate), and 62.9\%p (advanced), representing significant gains over zero-shot baselines. These findings highlight BankMathBench as a reliable benchmark for evaluating and advancing LLMs’ numerical reasoning in real-world banking scenarios. 
 \\ \newline  }

\begin{document}

\maketitleabstract

\section{Introduction}
The rapid evolution of mobile banking has transformed how customers manage core financial products such as deposits, savings accounts, and loans. This technological shift has fueled the widespread adoption of digital banking services, enabling customers to complete transactions without visiting physical bank branches. In particular, large language model (LLM)-based chatbots have emerged as a core channel for digital customer service by providing real-time responses to inquiries, reducing the workload of human agents, and improving overall accessibility and convenience \citep{takayanagi2025finpersona, yan2025creditllm, lakkaraju2023llms}.

Recent research has aimed to advance the mathematical reasoning capabilities of LLMs through specialized adaptation techniques and large-scale synthetic data generation \citep{yang2024qwen2, shao2024deepseekmath, chen2025advancing}. However, LLMs still exhibit clear limitations when confronted with complex numerical problems \citep{li-etal-2025-exposing, zhao-etal-2024-docmath, zheng-etal-2025-processbench}, particularly in real-world banking scenarios. This challenge stems from the fact that banking-related reasoning requires not only precise computation but also interpretive understanding of product-specific conditions \citep{srivastava-etal-2024-evaluating, tang-etal-2025-financereasoning}. Such errors or inconsistencies can directly undermine the trustworthiness of financial institutions, where reliability and factual correctness are paramount.

\begin{table}[t]
\renewcommand{\arraystretch}{1.5} 
\resizebox{\columnwidth}{!}{%
\begin{tabular}{lccc} 
\noalign{\hrule height 1pt}
\textbf{\large Model} & \textbf{\large Basic} & \textbf{\large Interm.} &  \textbf{\large Adv.} \\
\noalign{\hrule height 0.6pt}

\multicolumn{4}{l}{\textbf{\large Financial LLMs}} \\
{\large AdaptLLM} \citep{cheng2024adapting} & \large 0.4 & \large 0.0 & \large 0.0 \\
{\large Fin-o1} \citep{qian2025fino1} & \large 3.2 & \large 3.3 & \large 0.3 \\
\noalign{\hrule height 0.8pt}

\multicolumn{4}{l}{\textbf{\large Math-specialized LLMs}} \\
{\large DeepSeek-Math-Inst.} \citep{shao2024deepseekmath} & \large 50.4 & \large 1.4 & \large 0.1 \\
{\large Qwen2.5-Math-Inst.} \citep{yang2024qwen2} & \large 61.2 & \large 5.0 & \large 2.5 \\
\noalign{\hrule height 0.8pt}

\multicolumn{4}{l}{\textbf{\large Closed-source Model}} \\
{\large Claude Sonnet-3.5} \citep{Claude3S} & \large 66.7 & \large 9.9 & \large 10.0 \\
{\large Gemini-2.5-flash} \citep{comanici2025gemini} & \large 71.2 & \large 11.8 & \large 10.3 \\
{\large GPT-4o} \citep{hurst2024gpt} & \large 67.8 & \large 14.4 & \large 6.3 \\
\noalign{\hrule height 1pt}
\end{tabular}%
}
\caption{Zero-shot accuracy (\%) of various models across the Basic, Intermediate (Interm.), and Advanced (Adv.) difficulty levels defined in the BankMathBench dataset. Inst. denotes instruction-tuned models.}
\label{tab:zero_shot}
\end{table}

\begin{figure}[t] 
\centering
\includegraphics[width=\columnwidth]{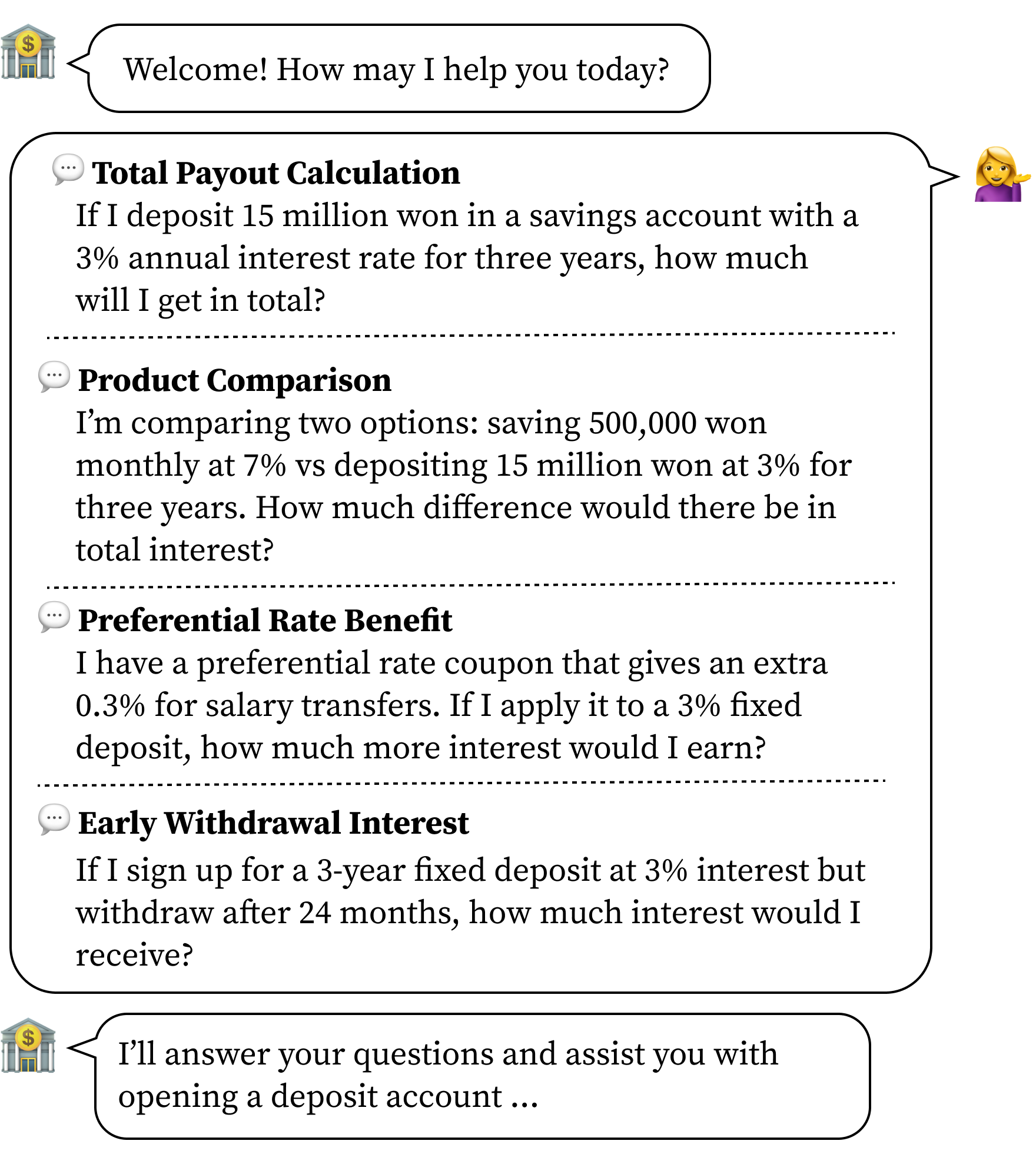} 
\caption{Examples of frequently asked customer queries in real banking branches.}
\label{fig:example}
\end{figure}

Existing mathematical benchmarks primarily consist of problems drawn from educational curricula (e.g., arithmetic or algebra), and thus fail to capture the domain-specific characteristics of numerical reasoning in practical banking contexts \citep{cobbe2021training, hendrycksmath2021}. Meanwhile, financial numerical reasoning benchmarks have largely focused on financial statements and securities analysis, leaving banking scenarios—such as deposits, savings, and loans—underexplored \citep{chen-etal-2021-finqa, chen-etal-2022-convfinqa, zhu-etal-2021-tat, zhao-etal-2024-knowledgefmath}. As shown in Table~\ref{tab:zero_shot}, both math-specialized and financial LLMs—trained on mathematical reasoning and financial numerical benchmarks, respectively—perform poorly in practical banking scenarios. Furthermore, even closed-source models show a marked performance drop with increasing task complexity, confirming that existing benchmarks generalize poorly to realistic banking scenarios. These results suggest that previous benchmarks have overlooked customer-critical queries requiring condition-specific computations, 
such as early-withdrawal interest, preferential-rate payouts, and product-level comparisons, as illustrated in Figure \ref{fig:example}.

To bridge this gap, we introduce BankMathBench, a domain-specific dataset for everyday banking scenarios. It consists of question–answer–reasoning triplets generated through a fully automated pipeline, which systematically constructs data across multiple levels of difficulty— from numerical reasoning on representative banking products to comparative reasoning across multiple products, and extending to multi-condition–based reasoning. As the first comprehensive benchmark for numerical reasoning in everyday banking, BankMathBench provides a reliable foundation for evaluating and improving LLM performance in this domain. Leveraging this high-quality dataset, we fine-tune open-source LLMs and demonstrate significant improvements in the accuracy of numerical reasoning in banking scenarios. Furthermore, by integrating the fine-tuned LLM’s formula generation capability with tool augmentation, we achieve substantial and consistent performance gains across diverse reasoning categories, highlighting the effectiveness of our proposed approach.

\section{Related Works}
\subsection{Numerical Reasoning with Large Language Models}
Recent studies have placed increasing emphasis on improving the numerical reasoning capabilities of LLMs for mathematical and financial applications. Some studies have aimed to improve the reasoning process itself. For example, techniques such as masking reasoning steps to guide learning \citep{chen-etal-2024-masked}, designing fine-grained reward schemes \citep{hwang-etal-2024-self}, separating computation from reasoning through program-based approaches \citep{chen2023program}, and decomposing problems into step-by-step subproblems \citep{imani-etal-2023-mathprompter} have been proposed. Another line of research integrates external tools such as Python to enhance the computational accuracy of LLMs \citep{zhang2023evaluating, wang2024mathcoder, yin-etal-2024-mumath}. Furthermore, OpenMath-CodeLlama \citep{NEURIPS2024_3d5aa9a7}, MetaMath \citep{yu2024metamath}, DeepSeekMath \citep{shao2024deepseekmath}, and Qwen2.5-Math \citep{yang2024qwen2} employ large-scale mathematical reasoning datasets to systematically improve their computational reasoning abilities. However, these studies predominantly evaluate performance on academic benchmarks, including GSM8K \citep{cobbe2021training} and MATH \citep{hendrycksmath2021}, thereby limiting their applicability to the financial domain.

A few studies have focused specifically on numerical reasoning within the financial domain. FinMath \citep{li2022finmath} designed a tree-structured solver to handle table–text hybrid financial reports, while APOLLO \citep{sun-etal-2024-apollo} proposed a method to generate executable program-based numerical reasoning for long-form documents. Nevertheless, these approaches do not fully capture the domain-specific reasoning required for everyday banking operations. Unlike previous studies, our work enables the assessment and improvement of LLMs’ numerical reasoning skills in practical banking scenarios.

\begin{figure*}[t]
\centering
{\phantomsubcaption\label{fig:main_framework-a}
\phantomsubcaption\label{fig:main_framework-b}
\phantomsubcaption\label{fig:main_framework-c}}
\includegraphics[width=\textwidth]{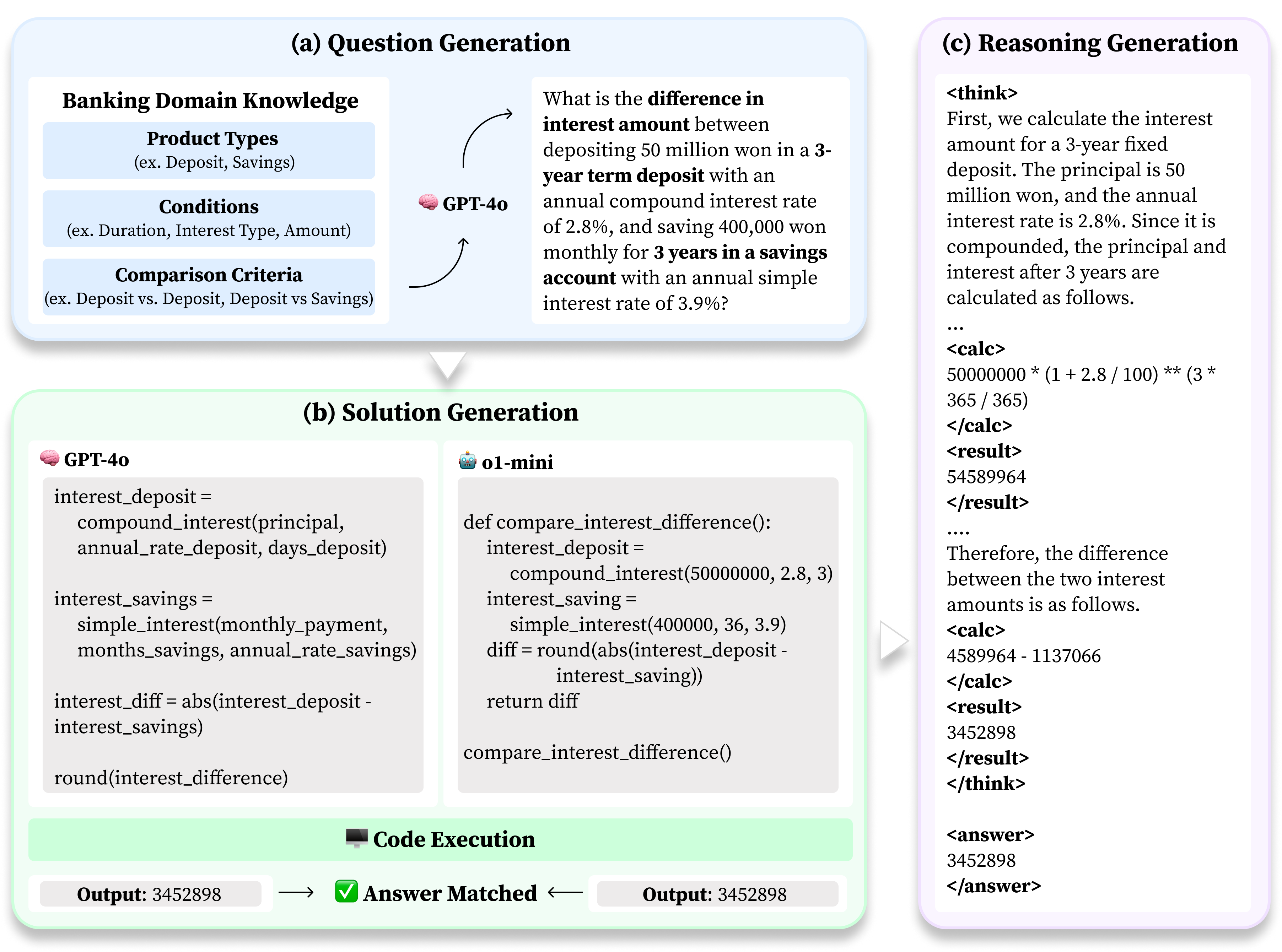}
\caption{Overview of the BankMathBench data generation pipeline, which comprises three stages: (a) question generation, (b) solution generation and automatic verification, and (c) reasoning generation.}
\label{fig:main_framework}
\end{figure*}

\subsection{Numerical reasoning Benchmarks in Financial Domain}
Specialized benchmarks are increasingly being introduced to evaluate numerical reasoning within the context of finance. FinQA \citep{chen-etal-2021-finqa}, and TAT-QA \citep{zhu-etal-2021-tat} primarily focus on interpreting financial reports and statements, while FinanceMath \citep{zhao-etal-2024-knowledgefmath} evaluates financial reasoning using college-level finance and economics problems. More recently, benchmarks such as FinReasoning \citep{tang-etal-2025-financereasoning}, Fin-R1 \citep{liu2025fin}, and XFinBench \citep{zhang-etal-2025-xfinbench} have been introduced to expand coverage and difficulty, and AlphaFin \citep{li-etal-2024-alphafin} extends the scope to corporate analysis and stock prediction. However, these benchmarks primarily focus on financial tasks such as investment, securities trading, or certification exams—while overlooking the fine-grained numerical reasoning required in everyday banking tasks. To fill this void, we introduce a rigorously curated and expert-validated dataset encompassing multiple difficulty levels and equipped with a reproducible automated generation pipeline for extensibility.

\begin{table}[t]
\renewcommand{\arraystretch}{1.3}
\centering
\resizebox{\columnwidth}{!}{%
{\footnotesize 
\begin{tabular}{llrr}
\noalign{\hrule height 1pt}
\textbf{Level} & \textbf{Language} & \textbf{Train Set} & \textbf{Test Set} \\
\noalign{\hrule height 0.6pt}

Basic & Korean  & 2,236 & 559 \\
      & English & 2,236 & 559 \\[2pt] \hline
Intermediate & Korean  & 1,708 & 426 \\
             & English & 1,702 & 423 \\[2pt] \hline
Advanced & Korean  & 1,596 & 399 \\ 
         & English & 1,596 & 399 \\
\noalign{\hrule height 0.8pt}
\textbf{Total} &  & \textbf{11,074} & \textbf{2,765} \\
\noalign{\hrule height 1pt}
\end{tabular}%
} 
}
\caption{Dataset statistics by difficulty level, showing the number of training and test samples for both Korean and English.}
\label{tab:num_banker_stats}
\end{table}

\begin{table*}[t]
\centering
\footnotesize 
\renewcommand{\arraystretch}{1.4}
\resizebox{\textwidth}{!}{%
\begin{tabular}{
  >{\centering\arraybackslash}m{2.2cm} |
  >{\raggedright\arraybackslash}p{0.44\textwidth} |
  >{\raggedright\arraybackslash}p{0.44\textwidth}}
\noalign{\hrule height 1pt}
\textbf{Level} & \textbf{Korean Question} & \textbf{English Question} \\
\noalign{\hrule height 0.6pt}

\textbf{Basic} &
매달 50만원씩 2년 동안 자유적금에 넣으면, 연 이자율이 복리 5.2\%일 때 만기 수령 금액은 얼마인가요? &
If you deposit 500{,}000 KRW every month into a free savings account for 2 years, what will be the maturity amount if the annual compound interest rate is 5.2\%? \\ 
\noalign{\hrule height 0.8pt}

\textbf{Intermediate} &
3년 만기 정기예금에 연 5.2\% 단리로 700만원을 넣는 것과, 매월 20만원씩 36개월 적금에 연 6.5\% 복리로 적금하는 것 중 만기 수령액 차이가 얼마나 되나요? &
What is the difference in the maturity amount between depositing 7 million won in a 3-year fixed deposit with an annual simple interest rate of 5.2\%, and saving 200{,}000 won every month for 36 months with an annual compound interest rate of 6.5\%? \\
\noalign{\hrule height 0.8pt}

\textbf{Advanced} &
B씨는 개인사업자대출로 1억 850만 원을 연 3.81\% 변동금리로 1년 동안 원금균등상환 방식으로 대출받았습니다. 6개월 후 금리가 0.3\%p 증가하여 연 4.11\%가 적용됩니다. 이 경우, 대출 기간 동안 내야 하는 이자는 얼마인가요? &
Mr. B took out a business loan of 108.5 million won with an annual variable interest rate of 3.81\% for one year, using the equal principal repayment method. After 6 months, the interest rate increased by 0.3 percentage points to 4.11\% annually. In this case, how much interest does he have to pay during the loan period? \\
\noalign{\hrule height 1pt}
\end{tabular}%
}
\caption{Representative questions from BankMathBench dataset demonstrating variations in difficulty levels.}
\label{tab:BankMathBench-examples}
\end{table*}

\section{Methodology}

This section introduces BankMathBench, a domain-specific dataset designed to systematically evaluate LLMs’ numerical reasoning capabilities in everyday banking scenarios. The dataset comprises three difficulty levels—basic, intermediate, and advanced—to enable comprehensive assessment across reasoning complexity. Specifically, the dataset encompasses the three core categories of banking products—deposits, savings, and loans. Problems involving deposits and savings require computing interest or total payout based on principal, interest type (simple or compound), frequency, and period. Loan-related problems include various loan types (e.g., mortgage, credit, automobile, and lease deposit loans) and repayment methods (e.g., equal principal and interest, or equal principal repayments). 

BankMathBench consists of 13,839 problems in both Korean and English (Table~\ref{tab:num_banker_stats}), each comprising a question, step-by-step reasoning, and final answer. The reasoning component guides LLMs in formulating correct equations and performing accurate calculations, ensuring that the relevant banking context and logical consistency are faithfully represented. The dataset was generated through an automated pipeline employing GPT-4o (version 2024-05-13) \citep{hurst2024gpt} and o1-mini (version 2024-09-12) \citep{jaech2024openai}, followed by manual verification by banking professionals. The parameterized data generation pipeline allows key variables—such as interest rates, tax policies, and currency units—to be adjusted, enhancing the portability of the dataset across national financial contexts.
Figure~\ref{fig:main_framework} and Table~\ref{tab:BankMathBench-examples} illustrate the pipeline and example instances, and the detailed prompts used in the generation process are provided in Appendix~\ref{appendix_a}. 

\subsection{Question Generation}
Questions are stratified by difficulty, based on the complexity of formulas. They are generated using GPT-4o with carefully designed prompt templates to produce realistic and computable problems, specifying product types and conditions, including clear comparison criteria for multi-product scenarios, as shown in Figure~\ref{fig:main_framework-a}. By default, numerical values such as amounts, interest rates, durations, and interest types are diversified to improve realism and coverage.

\subsubsection{Basic Level}
Basic-level questions cover actual banking products such as deposits, savings, and loans. These questions require multi-step calculations, including arithmetic, arithmetic/geometric progressions, and decimal operations. Each question specifies all product conditions—amount, interest rate, duration, payment frequency, and interest calculation method—to guarantee numerical solvability.

\subsubsection{Intermediate Level}
Intermediate-level questions involve multi-product comparisons, generating problems that consider two products simultaneously (deposit, savings, or loan). These questions are more challenging, as they require performing separate computations for each product and subsequently comparing the results—thereby increasing computational complexity and necessitating concurrent reasoning over conditions and the sequence of operations. Product pairings are randomly sampled from realistic combinations (e.g., deposit vs deposit, deposit vs savings, savings vs savings, loan vs loan), while unrealistic pairings (e.g., deposit vs loan) are excluded. Each question explicitly specifies the comparison criterion—whether the evaluation is based on interest rates or total payouts.

\subsubsection{Advanced Level}
Advanced-level questions extend single-product tasks to include complex multi-condition scenarios. Each product type follows a structured schema with required and optional fields: deposits and savings include taxation, early withdrawal options, and user-specific conditions (e.g., preferential rates based on deposit amount); loans include principal/interest repayment, interest type (fixed/variable), variable interest period, and prepayment fees. Feasible calculations are guaranteed by capping early withdrawal rates at the nominal annual rate and modeling variable loans with interest type, fluctuation period, and maximum residual-term change.

\subsubsection{Data Augmentation}
To enhance the diversity of questions, we applied a numerical augmentation process. This was motivated by the observation that LLMs tend to perform well with clean, round numbers (e.g., 2.5 or 10) but struggle with more realistic, irregular values (e.g., 2.93 or 3.1), which are common in real banking scenarios. 

\begin{figure}[h!] 
\centering
\includegraphics[width=\columnwidth]{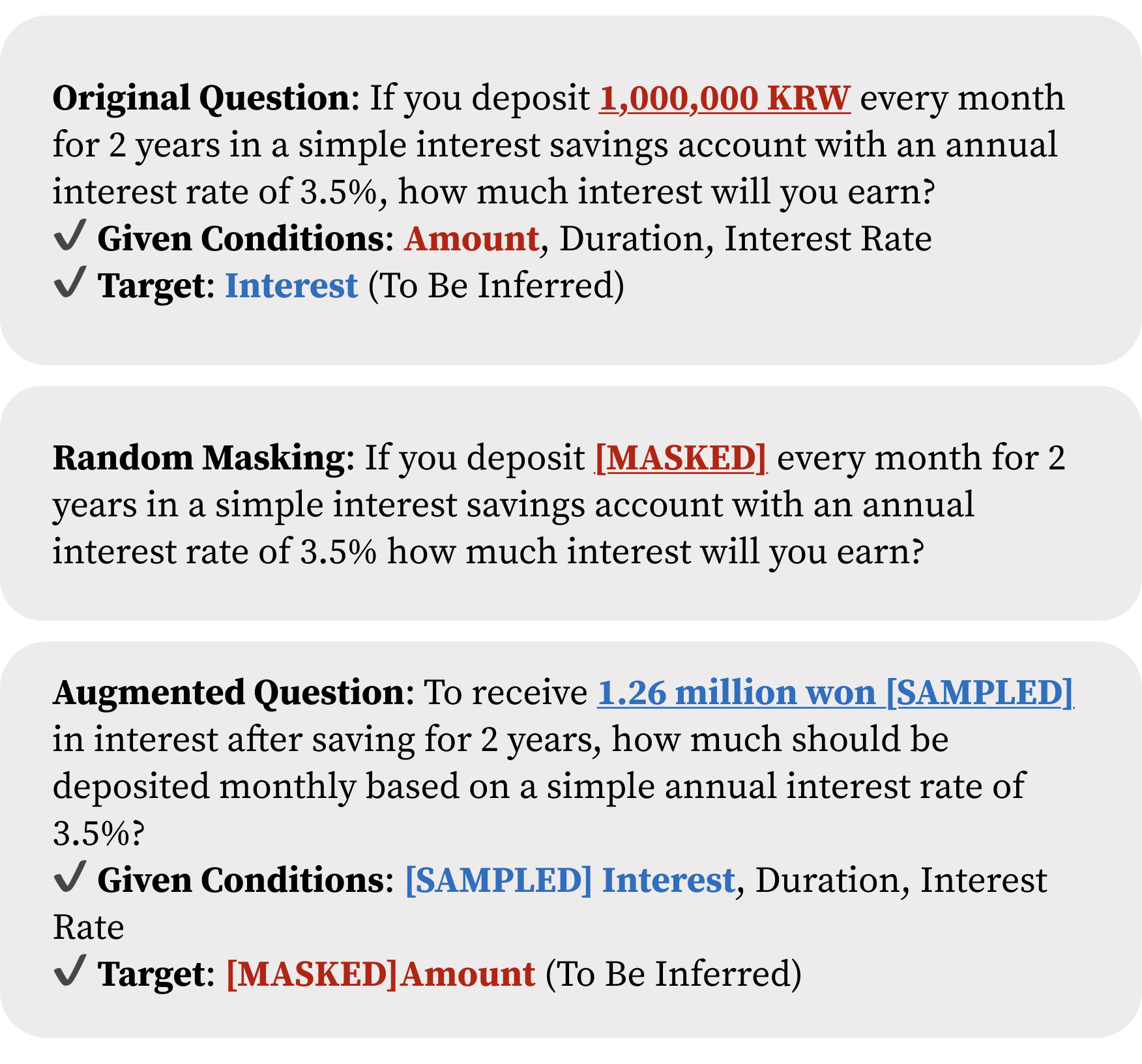} 
\caption{Example of the question augmentation process.}
\label{fig:augmentation}
\end{figure}

Additionally, we introduced random masking for basic-level questions by omitting certain numeric conditions, creating variants that require inferring the missing values. Essential numerical conditions for the augmented questions (e.g., 1.26 million won in Figure~\ref{fig:augmentation}) were randomly inserted during this process. This approach encourages models to perform contextual reasoning, interpreting surrounding information to derive the missing numerical conditions. Questions generated through this augmentation account for approximately 47\% of the intermediate dataset.

\subsection{Solution Generation and Data Filtering}
To obtain reliable and accurate answers, we applied a two-tiered solution generation strategy by problem difficulty. The approach guarantees automated verification of correctness through two distinct solutions for the same problem, with all calculations expressed in executable formats.
For basic-level questions, GPT-4o generates two fully computable formula representations—mathematical expressions and LaTeX. For intermediate and advanced-level questions involving multiple conditions or products, models (GPT-4o, o1-mini) generate structured, executable Python code. Python modules then compute answers, and only pairs with consistent results across formulas are retained, enabling large-scale automatic verification with minimal human effort (Figure \ref{fig:main_framework-b}).

\subsection{Reasoning Generation}
For each problem, GPT-4o is employed to generate stepwise reasoning from the question, solution, and answer. Given a question–solution pair as input, the model generates reasoning that integrates natural language and mathematical expressions. Reasoning steps are captured using <think> and </think> tags, and final answers are stored within <answer> and </answer> tags. To effectively leverage the LLM’s formula-generation capabilities and simultaneously prevent error accumulation, computable expressions are enclosed within <calc> and </calc> tags, with their results recorded in <result> and </result> tags, following prior work \citep{kadlcik-etal-2023-calc, goyal2024think}. Its computable, structured format allows precise tracking of stepwise computations and can be combined with external tools to enhance accuracy. Ultimately, it produces reasoning data, as illustrated in Figure \ref{fig:main_framework-c}, resulting in a complete question–answer–reasoning triplet.

\subsection{Expert Validation}
All question–answer–reasoning triplets were reviewed by two banking professionals with over three years of experience in retail banking to ensure the validity and reliability of the dataset. During this process, any identified inaccuracies or inconsistencies in problem statements, calculations, or reasoning steps were corrected. The evaluation focused on practical relevance and the logical correctness of calculations in real-world banking service contexts, based on the review criteria described below:

\begin{itemize}
\item Practical Relevance — Does the question reflect scenarios that could arise in actual bank consultations or customer interactions?
\item Logical Soundness — Are the solution and calculation procedures logically valid, considering the characteristics and rules of banking products?
\item Format Consistency — Are the data structure, formula representations, units, and notations consistent across the entire dataset?
\end{itemize}

Through this review, calculation errors and ambiguous expressions are eliminated, so that the dataset achieves a level of financial reasoning quality suitable for real-world banking applications.

\section{Experimental Design}
We conduct experiments on a range of LLMs with up to 8 billion parameters to evaluate their numerical reasoning capabilities on the BankMathBench benchmark. 
The evaluated models include open-source instruction-tuned variants such as LLaMA-3 \citep{grattafiori2024llama}, DeepSeek-R1-Distill \citep{guo2025deepseek}, Qwen2.5-Instruct \citep{Yang2024Qwen25TR}, Phi-3.5-Mini-Instruct \citep{Abdin2024Phi3TR}, and Kanana \citep{bak2025kanana}, as well as reasoning-enhanced or math-specialized variants such as Qwen2.5-Math \citep{yang2024qwen2}, DeepSeek-Math \citep{shao2024deepseekmath}, and Qwen3 \citep{yang2025qwen3}. Additionally, the closed-source models—Claude Sonnet-3.5 (2024-06-20 version) \citep{Claude3S}, Gemini-2.5-Flash (2025-06-17 version) \citep{comanici2025gemini}, and GPT-4o (2024-05-13 version) \citep{hurst2024gpt}—are also included.

\begin{figure}[h] 
\centering
\includegraphics[width=\columnwidth]{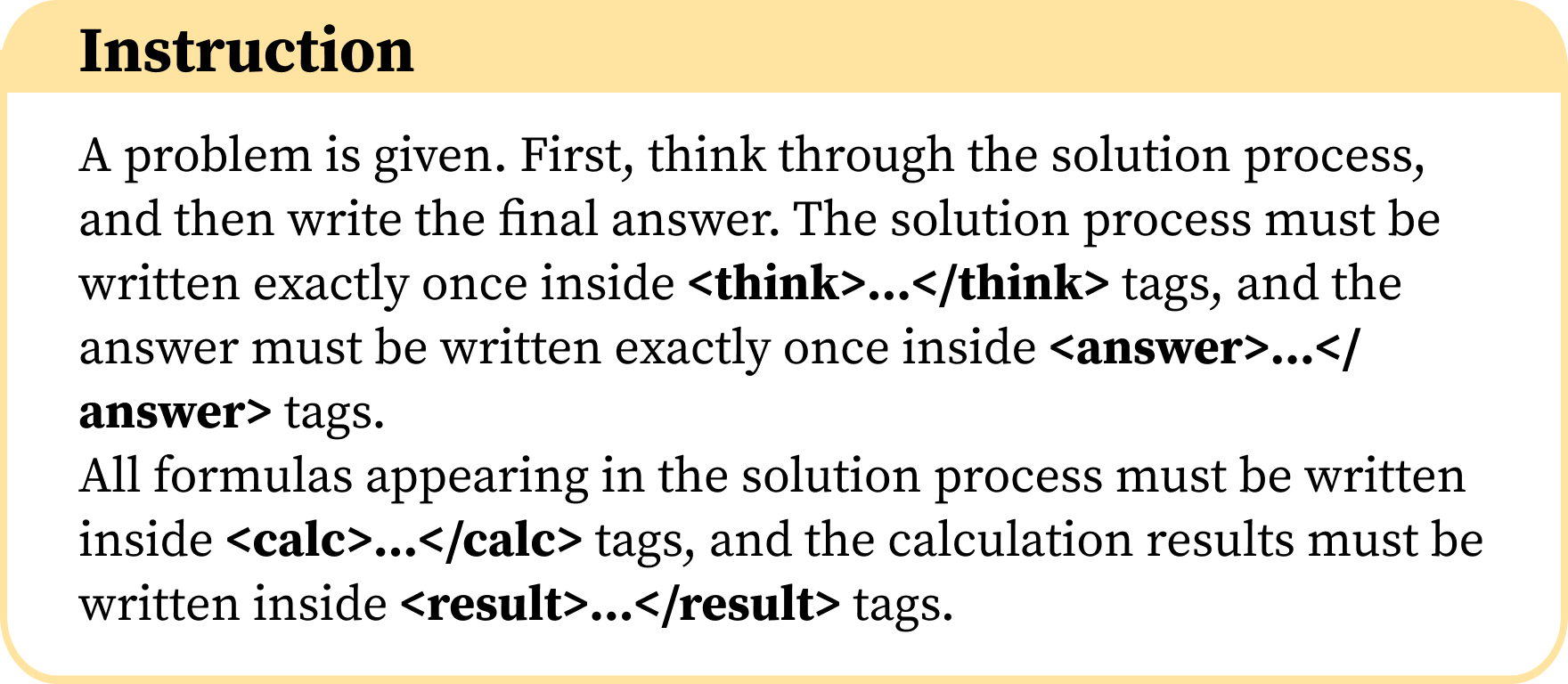} 
\caption{Instruction format used across all models for zero-shot evaluation and supervised fine-tuning (SFT).}
\label{fig:instruction}
\end{figure}

The BankMathBench dataset was divided into training and test sets (8:2 split).
Evaluation was based on exact-match accuracy (pass@1), with normalization applied to equivalent numerical or currency expressions (e.g., “₩10 million” and “10,000,000 KRW”).
For zero-shot evaluations, multiple valid output formats (e.g., \textbackslash boxed\{\}, <answer>...</answer>) were accepted, while fine-tuned models were required to adhere strictly to the instructed response format. All models followed the same instruction format, as shown in Figure \ref{fig:instruction}.

SFT was conducted using 4-bit quantized LoRA (QLoRA) \citep{dettmers2023qlora} with rank $r=16$, scaling factor $\alpha=16$, learning rate $2\times10^{-4}$, and batch size 4 for 10 epochs on four NVIDIA RTX 3090 GPUs. To further assess improvements in numerical precision, a tool-augmented SFT was employed, where equations enclosed within <calc>...</calc> tags were automatically handled by tool calling to an external calculator, and the resulting values were inserted sequentially into <result>...</result> tags.

\begin{table}[t]
\renewcommand{\arraystretch}{1.2}
\centering
\resizebox{\columnwidth}{!}{%
\begin{tabular}{l|c|cc|cc|cc}
\noalign{\hrule height 1.2pt}
\multicolumn{2}{c|}{\textbf{Model Info}} 
& \multicolumn{2}{c|}{\textbf{Zero-shot}} 
& \multicolumn{2}{c|}{\textbf{SFT}} 
& \multicolumn{2}{c}{\textbf{Tool Aug. SFT}} \\ \hline
\textbf{Model} & \textbf{Size} & \textbf{Kor} & \textbf{Eng} & \textbf{Kor} & \textbf{Eng} & \textbf{Kor} & \textbf{Eng} \\ \hline
Claude Sonnet-3.5 & - & 69.2 & 66.7 & - & - & - & - \\ \hline
Gemini-2.5-Flash  & - & 64.9 & 71.2 & - & - & - & - \\ \hline
GPT-4o & - & 67.8 & 67.8 & - & - & - & - \\ \hline
\noalign{\hrule height 0.5pt}
DeepSeek-R1-Distill-Qwen & 1.5B & 2.1 & 31.8 & 59.4 & 58.5 & 78.9 & 77.8 \\ \hline
DeepSeek-R1-Distill-Qwen & 7B & 26.1 & 26.1 & 58.9 & 64.2 & 79.6 & \cellcolor{top3}83.5 \\ \hline
DeepSeek-Math-Instruct & 7B & 27.9 & 50.4 & 57.2 & 64.9 & 69.6 & \cellcolor{top2}84.4 \\ \hline
Kanana-nano-Base & 2.1B & 3.0 & 0.9 & 44.0 & 43.8 & 76.4 & 75.0 \\ \hline
Kanana-nano-Instruct & 2.1B & 10.7 & 26.7 & 44.4 & 46.0 & 72.1 & 76.2 \\ \hline
Kanana-1.5-Base & 2.1B & 0.7 & 7.9 & 47.6 & 49.6 & 75.0 & 74.8 \\ \hline
Kanana-1.5-Instruct & 2.1B & 9.5 & 21.1 & 41.3 & 45.8 & 68.9 & 78.9 \\ \hline
Kanana-1.5-Instruct & 8B & 25.8 & 40.8 & 51.0 & 51.7 & 74.4 & 76.7 \\ \hline
Llama-3.2-Instruct & 1B & 0.0 & 26.1 & 34.2 & 36.3 & 56.5 & 63.3 \\ \hline
Llama-3.2-Instruct & 3B & 2.9 & 44.5 & 52.4 & 53.8 & 79.6 & 78.0 \\ \hline
Llama-3-Instruct & 8B & 14.7 & 34.0 & 58.3 & 57.6 & 77.6 & 79.4 \\ \hline
Phi-3.5-Mini-Instruct & 3.8B & 10.0 & 49.6 & 54.2 & 57.8 & 68.9 & 80.5 \\ \hline
Qwen2.5-Instruct & 1.5B & 4.8 & 52.2 & 51.5 & 45.8 & 75.5 & 75.1 \\ \hline
Qwen2.5-Instruct & 3B & 24.2 & 56.4 & 59.6 & 53.3 & 77.5 & 80.7 \\ \hline
Qwen2.5-Instruct & 7B & 44.4 & 60.5 & 65.1 & 60.6 & \cellcolor{top3}83.0 & \cellcolor{top3}83.5 \\ \hline
Qwen2.5-Math-Instruct & 7B & 46.2 & 61.2 & 64.2 & 49.9 & \cellcolor{top2}83.4 & 74.6 \\ \hline
Qwen3 & 4B & 44.2 & 9.8 & 66.0 & 60.8 & 79.3 & 72.5 \\ \hline
Qwen3 & 8B & 27.2 & 21.6 & 69.2 & 66.2 & \cellcolor{top1}85.2 & \cellcolor{top1}85.2 \\
\noalign{\hrule height 1.2pt}
\end{tabular}%
}
\caption{Results on the BankMathBench Basic dataset. Model accuracy (\%) is reported for Zero-shot, SFT, and Tool-Augmented SFT settings across Korean (Kor) and English (Eng).}
\label{tab:v1_basic}
\end{table}

\section{Results}

\subsection{Overall Performance}
The zero-shot evaluation on the BankMathBench dataset revealed that model performance declined as dataset difficulty increased in both English and Korean. In particular, for the Korean dataset, open-source models achieved average accuracies of 18.0\%, 0.8\%, and 0.5\% on the basic, intermediate, and advanced subsets, respectively, exhibiting a sharp decrease as the reasoning complexity increased. This trend underscores the escalating reasoning and computational demands inherent in higher difficulty levels. Representative error types are provided in Appendix \ref{appendix_b}, offering a qualitative analysis of common failure cases. Our dataset is carefully designed to support error correction through supervised training and tool augmentation, ultimately yielding substantial performance gains.

Fine-tuning on the BankMathBench dataset led to consistent performance improvements across all models. Notably, Qwen3-8B and DeepSeek-Math-Instruct-7B achieved strong and balanced performance across all language settings and dataset difficulty levels. Specifically, Qwen3-8B exhibited substantial gains over zero-shot performance, increasing by 42.0 and 44.6 percentage points in Korean and English, respectively, on the basic-level dataset, while DeepSeek-Math-Instruct-7B improved by 29.3 and 14.5 percentage points (Table \ref{tab:v1_basic}). These results indicate that even math-oriented or reasoning-enhanced LLMs struggle with domain-specific numerical reasoning, and that fine-tuning on well-defined banking scenarios is essential for improving both reasoning capability and computational accuracy. 

\begin{table}[t]
\renewcommand{\arraystretch}{1.2}
\centering
\resizebox{\columnwidth}{!}{%
\begin{tabular}{l|c|cc|cc|cc}
\noalign{\hrule height 1.2pt}
\multicolumn{2}{c|}{\textbf{Model Info}} 
& \multicolumn{2}{c|}{\textbf{Zero-shot}} 
& \multicolumn{2}{c|}{\textbf{SFT}} 
& \multicolumn{2}{c}{\textbf{Tool Aug. SFT}} \\ \hline
\textbf{Model} & \textbf{Size} 
& \textbf{Kor} & \textbf{Eng} 
& \textbf{Kor} & \textbf{Eng} 
& \textbf{Kor} & \textbf{Eng} \\ 
\noalign{\hrule height 0.5pt}
Claude Sonnet-3.5 & - & 10.3 & 9.9 & - & - & - & - \\ \hline
Gemini-2.5-Flash  & - & 10.7 & 11.8 & - & - & - & - \\ \hline
GPT-4o & - & 7.9 & 14.4 & - & - & - & - \\ \hline
\noalign{\hrule height 0.5pt}
DeepSeek-R1-Distill-Qwen & 1.5B & 0.0 & 2.1 & 5.8 & 8.3 & 83.0 & 85.8 \\ \hline
DeepSeek-R1-Distill-Qwen & 7B & 1.6 & 0.2 & 1.9 & 7.3 & 69.2 & 88.7 \\ \hline
DeepSeek-Math-Instruct & 7B & 0.5 & 1.4 & 6.8 & 8.0 & \cellcolor{top2}86.5 & \cellcolor{top3}89.8 \\ \hline
Kanana-nano-Base & 2.1B & 0.0 & 0.0 & 2.1 & 2.4 & 81.9 & 85.3 \\ \hline
Kanana-nano-Instruct & 2.1B & 0.0 & 0.5 & 1.4 & 0.9 & 81.2 & 84.9 \\ \hline
Kanana-1.5-Base & 2.1B & 0.0 & 0.7 & 2.3 & 1.7 & 83.3 & 57.9 \\ \hline
Kanana-1.5-Instruct & 2.1B & 0.0 & 0.7 & 3.1 & 2.8 & 80.3 & 70.7 \\ \hline
Kanana-1.5-Instruct & 8B & 0.7 & 1.7 & 6.3 & 2.8 & \cellcolor{top3}85.9 & 61.9 \\ \hline
Llama-3.2-Instruct & 1B & 0.0 & 0.2 & 0.5 & 0.7 & 27.0 & 44.0 \\ \hline
Llama-3.2-Instruct & 3B & 0.5 & 3.5 & 4.5 & 4.5 & 72.3 & 79.4 \\ \hline
Llama-3-Instruct & 8B & 0.7 & 1.2 & 6.8 & 7.3 & 84.0 & 88.4 \\ \hline
Phi-3.5-Mini-Instruct & 3.8B & 0.5 & 6.4 & 4.9 & 4.5 & 61.0 & 80.4 \\ \hline
Qwen2.5-Instruct & 1.5B & 0.0 & 3.1 & 3.0 & 3.5 & 69.2 & 63.1 \\ \hline
Qwen2.5-Instruct & 3B & 0.2 & 5.4 & 6.3 & 6.9 & 81.6 & 86.8 \\ \hline
Qwen2.5-Instruct & 7B & 3.0 & 9.0 & 9.1 & 9.0 & \cellcolor{top1}90.9 & \cellcolor{top2}90.8 \\ \hline
Qwen2.5-Math-Instruct & 7B & 0.9 & 5.0 & 6.3 & 6.1 & 81.1 & 87.5 \\ \hline
Qwen3 & 4B & 2.6 & 6.6 & 5.6 & 7.3 & 61.8 & 64.3 \\ \hline
Qwen3 & 8B & 2.8 & 5.0 & 9.1 & 9.2 & 85.8 & \cellcolor{top1}92.7 \\ 
\noalign{\hrule height 1.2pt}
\end{tabular}%
}
\caption{Results on the BankMathBench Intermediate dataset. Model accuracy (\%) is reported for Zero-shot, SFT, and Tool-Augmented SFT settings across Korean (Kor) and English (Eng).}
\label{tab:v2_intermediate}
\end{table}

Building upon these results, tool-augmented supervised fine-tuning further strengthened the models’ numerical reasoning ability. 
When compared with SFT, tool-augmentation fine-tuning yielded significant gains across datasets of varying difficulty and languages, increasing performance by 21.3, 71.1, and 62.8 percentage points for the basic, intermediate, and advanced Korean datasets, and by 24.1, 72.7, and 63.4 percentage points for the corresponding English datasets, respectively.
These findings suggest that our dataset reinforces the model’s intrinsic reasoning foundation, enabling tool-augmented reasoning to yield substantial performance gains in banking scenarios. In particular, the intermediate level—which requires generating multiple formulas for each product—exhibited the largest performance gains after applying external tools. This indicates that while SFT effectively improved formula generation, models still struggled to aggregate multiple computed results into a single correct outcome. The introduction of tool-augmented fine-tuning alleviated this reasoning bottleneck by enabling more accurate multi-step computation and integration across intermediate-level tasks.

\begin{table}[t]
\renewcommand{\arraystretch}{1.2}
\centering
\resizebox{\columnwidth}{!}{%
\begin{tabular}{l|c|cc|cc|cc}
\noalign{\hrule height 1.2pt}
\multicolumn{2}{c|}{\textbf{Model Info}} 
& \multicolumn{2}{c|}{\textbf{Zero-shot}} 
& \multicolumn{2}{c|}{\textbf{SFT}} 
& \multicolumn{2}{c}{\textbf{Tool Aug. SFT}} \\ \hline
\textbf{Model} & \textbf{Size} 
& \textbf{Kor} & \textbf{Eng} 
& \textbf{Kor} & \textbf{Eng} 
& \textbf{Kor} & \textbf{Eng} \\ 
\noalign{\hrule height 0.6pt}
Claude Sonnet-3.5 & - & 15.8 & 10.0 & - & - & - & - \\ \hline
Gemini-2.5-Flash  & - & 10.3 & 10.3 & - & - & - & - \\ \hline 
GPT-4o & - & 4.3 & 6.3 & - & - & - & - \\ \hline
\noalign{\hrule height 0.5pt}
DeepSeek-R1-Distill-Qwen & 1.5B & 0.0 & 0.3 & 0.3 & 0.3 & 58.4 & 69.4 \\ \hline
DeepSeek-R1-Distill-Qwen & 7B & 1.0 & 0.0 & 2.0 & 2.0 & 68.9 & 72.4 \\ \hline
DeepSeek-Math-Instruct & 7B & 0.5 & 1.0 & 1.5 & 0.8 & \cellcolor{top3}71.7 & \cellcolor{top1}77.2 \\ \hline
Kanana-nano-Base & 2.1B & 0.0 & 0.0 & 0.0 & 0.0 & 65.7 & 62.2 \\ \hline
Kanana-nano-Instruct & 2.1B & 0.0 & 0.0 & 0.3 & 0.0 & 59.6 & 56.6 \\ \hline
Kanana-1.5-Base & 2.1B & 0.0 & 0.0 & 0.0 & 0.0 & 62.4 & 51.6 \\ \hline
Kanana-1.5-Instruct & 2.1B & 0.0 & 0.0 & 0.0 & 0.0 & 64.9 & 59.9 \\ \hline
Kanana-1.5-Instruct & 8B & 0.0 & 0.0 & 0.0 & 0.0 & 66.9 & 59.1 \\ \hline
Llama-3.2-Instruct & 1B & 0.0 & 0.0 & 0.0 & 0.0 & 36.6 & 33.1 \\ \hline
Llama-3.2-Instruct & 3B & 0.0 & 0.3 & 0.0 & 0.0 & 64.4 & 65.4 \\ \hline
Llama-3-Instruct & 8B & 0.0 & 0.5 & 0.0 & 0.3 & 69.7 & 68.4 \\ \hline
Phi-3.5-Mini-Instruct & 3.8B & 0.0 & 1.0 & 0.3 & 0.5 & 41.4 & 59.9 \\ \hline
Qwen2.5-Instruct & 1.5B & 0.0 & 0.3 & 0.3 & 0.0 & 40.4 & 59.4 \\ \hline
Qwen2.5-Instruct & 3B & 0.8 & 0.5 & 0.0 & 0.0 & \cellcolor{top3}71.7 & 70.2 \\ \hline
Qwen2.5-Instruct & 7B & 1.0 & 2.8 & 1.0 & 0.5 & 70.9 & 69.9 \\ \hline
Qwen2.5-Math-Instruct & 7B & 2.0 & 2.5 & 1.3 & 1.3 & 70.7 & 65.9 \\ \hline
Qwen3 & 4B & 1.5 & 0.3 & 1.8 & 2.5 & \cellcolor{top2}76.9 & \cellcolor{top2} 75.7 \\ \hline
Qwen3 & 8B & 2.8 & 0.5 & 1.8 & 1.8 & \cellcolor{top1}80.0 & \cellcolor{top3} 74.0 \\ 
\noalign{\hrule height 1.2pt}
\end{tabular}%
}
\caption{Results on the BankMathBench Advanced dataset. Model accuracy (\%) is reported for Zero-shot, SFT, and Tool-Augmented SFT settings across Korean (Kor) and English (Eng).}
\label{tab:v2_advanced}
\end{table}

Further analysis indicated that the influence of training data language on model performance became more pronounced as dataset difficulty increased. Korean-specialized models such as the Kanana series achieved superior results on the Korean dataset, whereas other multilingual models performed relatively consistently across languages.

\subsection{Effect of Model Size and Architecture}

Across all datasets (Tables \ref{tab:v1_basic}-\ref{tab:v2_advanced}), larger backbone models (7B/8B) generally achieved the highest performance, confirming the expected correlation between model scale and reasoning capability. However, several notable exceptions were observed. The DeepSeek-R1-Distill-Qwen-1.5B model performed comparably to math-specialized backbones such as DeepSeek-Math-7B and Qwen-Math-7B, suggesting that lightweight architectures can still capture domain-specific numerical reasoning when fine-tuned effectively. Similarly, Kanana-nano-Base-2.1B exhibited performance on par with—or even surpassing—that of the larger Kanana-1.5-Instruct-8B model. Interestingly, LLaMA-3.2-Instruct-1B, which lacks official Korean language support, exhibited very low performance; however, scaling up to the 3B configuration alone led to considerable improvements across all benchmarks.

Despite these architectural and scale differences, fine-tuning led to more stable performance across models. Additionally, empirical results obtained from the Korean basic-level dataset revealed high variance in the zero-shot setting ($\mu$ = 18.0, $\sigma$ = 15.8), which was markedly reduced after fine-tuning ($\mu$ = 54.4, $\sigma$ = 9.5), a trend that was also consistent across the English dataset. These results suggest consistent convergence across different model sizes on our dataset and demonstrate that domain-specific fine-tuning effectively stabilizes performance even when updating only a subset of parameters.

\subsection{Impact of Fine-Tuning on Numerical Reasoning Capability}

To assess the impact of fine-tuning on numerical reasoning, we conducted an error analysis of the Qwen3-8B model trained on the BankMathBench advanced dataset. Specifically, we measured the absolute differences and relative error ratios between predicted and ground-truth values. To account for the large variability in numerical values, a log-scale transformation (base 1000) was applied, and median-based comparisons were used. 

\begin{table}[h]
\renewcommand{\arraystretch}{1.4}
\centering
\resizebox{\columnwidth}{!}{%
{\small
\begin{tabular}{lrrrrrr}
\noalign{\hrule height 1pt}

\textbf{} & \multicolumn{2}{c}{\textbf{Absolute Error $\downarrow$ }} & \multicolumn{2}{c}{\textbf{Error Ratio $\downarrow$}} & \multicolumn{2}{c}{\textbf{Accuracy $\uparrow$}} \\
\noalign{\hrule height 0.6pt}
 & \textbf{Kor} & \textbf{Eng} & \textbf{Kor} & \textbf{Eng} & \textbf{Kor} & \textbf{Eng} \\
\noalign{\hrule height 0.6pt}

\textbf{Zero-shot} & 2.65 & 2.65 & 100.00 & 100.00 & 2.75 & 0.50 \\ 
\textbf{SFT} & 2.05 & 2.12 & 1.28 & 1.89 & 1.75 & 1.75 \\ 
\textbf{Tool Aug. SFT} & 0.00 & 0.00 & 0.00 & 0.00 & 79.95 & 73.94 \\
\noalign{\hrule height 0.8pt}
\end{tabular}%
}
}
\caption{Comparison of absolute error, error ratio, and accuracy across Zero-shot, SFT, and Tool-Augmented SFT settings on the BankMathBench Advanced dataset.}
\label{tab:error_analysis}
\end{table}

As shown in Table~\ref{tab:error_analysis}, fine-tuning substantially reduced numerical prediction errors compared to the zero-shot baseline, indicating improved numerical precision and robustness even without external tool integration. Notably, while the model’s exact computation accuracy reached 1.75\% after fine-tuning, the median relative error ratios were only 1.28\% in Korean and 1.89\% in English, decreasing to nearly zero when combined with tool augmentation. These results demonstrate that fine-tuning on BankMathBench not only enables models to reliably generate correct formulas but also establishes a solid foundation for tool-augmented reasoning, effectively mitigating cumulative numerical errors. Collectively, these findings validate BankMathBench as an effective dataset for improving LLMs’ internal numerical consistency and reasoning stability.

\section{Conclusion} 
We introduce BankMathBench, a domain-specific benchmark for banking that addresses a previously underrepresented area in numerical reasoning. The BankMathBench is developed to evaluate models’ financial calculation abilities across three levels of task difficulty in banking scenarios. Models fine-tuned on BankMathBench exhibited significant gains in formula generation and computational accuracy, evidencing the dataset’s effectiveness in strengthening numerical reasoning in LLMs. Notably, tool-augmented fine-tuning yielded greater improvements on complex reasoning tasks, as BankMathBench enhances models’ formula-generation capabilities, enabling them to leverage external tools effectively for sophisticated banking computations. We expect that BankMathBench will foster further research on developing banking-specialized reasoning models and inspire the exploration of tool-integrated learning approaches for more robust numerical reasoning. Ultimately, we envision BankMathBench-trained models enabling digital banking assistants that can provide accurate, computation-grounded responses tailored to customers’ personal contexts.

\section{Limitations}

While BankMathBench effectively captures numerical reasoning across key retail banking products, its current coverage is limited to representative scenarios involving deposits, savings, and loans. Other financial domains, such as funds and insurance, are not yet included, which may lead to product-specific biases in models fine-tuned on the dataset. Future work will expand the dataset’s scope by leveraging the proposed automated data generation pipeline to encompass a broader range of banking products and customer-oriented scenarios.

From an evaluation perspective, performance in this work was assessed primarily in terms of numerical correctness and error rates between predicted and ground-truth values. Improvements in accuracy may indirectly suggest enhanced reasoning quality; however, reasoning quality was not directly evaluated. Future research may incorporate explicit assessment of reasoning processes to provide a more comprehensive understanding of model performance in financial numerical reasoning tasks.

\section{Ethical Statements}
The BankMathBench dataset contains no harmful, offensive, or personally identifiable information. All data instances were synthetically generated using large language models and do not originate from customer records. The dataset was created independently and is free from any conflicts of interest.

Additionally, all human annotators and reviewers involved in the dataset verification process were fairly compensated, receiving payment above the minimum hourly wage (\$7) in South Korea in accordance with ethical research and labor standards.

\section{Data Availability}
The BankMathBench dataset will be made available in a public repository following completion of institutional review and compliance procedures. The approval process is currently in progress. Researchers interested in early access may contact the authors.

\section{Bibliographical References}

\bibliographystyle{lrec2026-natbib}
\bibliography{lrec2026-example}

\appendix
\section{Prompt Design}
\label{appendix_a}

The prompts for the BankMathBench data generation pipeline—question generation, solution generation, and reasoning generation—are presented in Tables~\ref{tab:basic_prompt}–\ref{tab:advanced_prompt}, Table~\ref{tab:solution_prompt}, and Table~\ref{tab:reasoning_prompt}, respectively. Across all stages, prompts are organized according to a common template comprising four core fields—role, requirements, example, and output format—which ensures structural consistency, domain alignment, and controllability. The fields are designed to support flexible adaptation across difficulty levels and banking product types, and their detailed contents (e.g., specific requirements or illustrative examples) can be modified or extended as needed.

\paragraph{Question Generation.} Difficulty-specific prompts are used to construct basic, intermediate, and advanced questions (Tables~\ref{tab:basic_prompt}–\ref{tab:advanced_prompt}). Each prompt encodes banking-domain constraints, including product types and financial conditions. In Table~\ref{tab:advanced_prompt}, bracketed placeholders (e.g., (Deposit | Savings), <Rate>) are randomly sampled and programmatically instantiated, and only the finalized values are propagated to the final prompt.

\paragraph{Solution Generation.} Ground-truth answers are obtained using the prompt in Table~\ref{tab:solution_prompt}. The prompt can be configured to produce solutions in standard mathematical form, LaTeX, or executable Python code, with reference equations adapted accordingly to preserve computational consistency.

\paragraph{Reasoning Generation.} Step-by-step reasoning traces are generated using the prompt in Table~\ref{tab:reasoning_prompt}. Given the question, the corresponding formula or code, and the verified final answer, the model produces structured reasoning with \texttt{<think></think>} and \texttt{<answer></answer>} tags, ensuring adherence to the specified format and requirements.

\section{Representative Error Types}
\label{appendix_b}
Table \ref{tab:error_type} summarizes representative error types in zero-shot financial reasoning. Type 1 (Formula Error) refers to applying an incorrect financial formula due to misunderstanding of the product structure. Type 2 (Parameter Conversion Error) captures mistakes in selecting or applying key parameters, such as using the wrong interest rate or misplacing tax deductions. Type 3 (Arithmetic Error) includes cases where the correct formula is used but the final numerical computation is incorrect. 

\begin{table*}[t]
\centering
\scriptsize
\setlength{\tabcolsep}{4pt}
\renewcommand{\arraystretch}{1.12}
\resizebox{\textwidth}{!}{
\begin{tabular}{p{0.985\textwidth}}
\noalign{\hrule height 1pt}

\textbf{(Kor)} \\
\textbf{[Role]} \\
당신은 현실에 기반한 상상력을 지닌 작가입니다. 대한민국 K은행을 방문하는 고객들의 실제 상담 장면을 바탕으로 소설을 쓰고 있습니다. 은행 상품의 기초 계산에 관심 있는 고객들의 페르소나와 질문을 생성하세요. 과장된 설정은 지양하고, 실제 상담 현장에서 있을 법한 자연스럽고 현실적인 상황을 기반으로 작성하세요. \\ \\

\textbf{[요구사항]} \\
- 고객의 상황이나 목표(은퇴 준비, 결혼 준비, 주택 구입, 자녀 교육자금 마련 등)를 함께 언급하세요. \\
- 질문은 ``상황 설명 + 구체적 계산 질문'' 형태로 작성하세요. \\
- 상품, 금액, 이자율, 기간, 납입주기, 이자 계산 방식 6가지 정보를 반드시 포함하세요. \\
- 상품 가입 조건을 고려하여 1$\sim$2문장으로 자연스럽고 간결한 질문을 작성합니다. \\
- 가입 조건이 맞지 않으면 질문을 생성하지 않습니다. \\
- 아래 예시와 동일한 내용은 생성하지 않습니다. \\ \\

\textbf{[예시]} \\
\# 은행 상품 예시: \\
정기예금, 정기적금, 파킹통장, 주택담보대출, 중고차담보대출, 전월세보증금대출, 개인사업자대출, 비상금대출 \\
\\[-1em]
\# 질문 예시: \\
- 곧 은퇴를 앞두고 있는 50대 입니다. 5억원을 정기예금 1년짜리(연 이자율 3.88\%) 단리식 상품에 넣으면 세후 이자가 얼마나 될까요? \\
- 취업한 지 6개월 된 사회 초년생입니다. 한 달에 30만 원씩 연 이율 7\% 복리식 자유적금에 저축하면 2년 뒤 결혼자금으로 얼마를 모을 수 있을까요? \\
- 현금으로 4억원 정도를 가지고 있어서 주택담보대출로 4억원을 받아서 아파트를 매매하려고 합니다. 연 이자율 2.1\%로 30년 원리금 균등상환으로 하면, 이자를 얼마씩 갚아야 하나요? \\ \\

\textbf{[출력 형식]} 생성된 질문을 다음 JSON 형식으로 출력하세요. \\
\texttt{\{}  
    \texttt{"상품": "\textless 상품 유형 (예: 예금, 적금, 대출)\textgreater", \textbackslash n } 
    \texttt{"질문": "\textless 생성된 질문\textgreater"} 
\texttt{\}} \\

\hline
\textbf{(Eng)} \\
\textbf{[Role]}
You are a writer with imaginative yet reality-grounded thinking. You are writing a novel based on real consultation scenes of customers visiting K Bank in South Korea. Generate customer personas and questions related to basic calculations of banking products. Avoid exaggerated scenarios and base your writing on realistic situations that could plausibly occur in actual bank consultations. \\ \\

\textbf{[Requirements]} \\
- Mention the customer’s situation or goal (e.g., retirement planning, wedding preparation, home purchase, saving for a child’s education). \\
- Write each question in the form: ``Situation description + specific calculation question.'' \\
- Each question must include six elements: product, amount, interest rate, term, payment frequency, and interest calculation method. \\
- Considering product eligibility conditions, write a natural and concise question within 1--2 sentences. \\
- Do not generate a question if eligibility conditions are not satisfied. \\
- Do not generate content identical to the examples below. \\ \\

\textbf{[Example]} \\

\# Example Bank Products: \\
Fixed deposit, installment savings, parking account, mortgage loan, used car secured loan, jeonse deposit loan, business loan, emergency loan \\
\\[-1em]
\# Example Questions:\\
- I am in my 50s and about to retire. If I deposit 500 million KRW into a 1-year fixed deposit (annual interest rate 3.88\%) with simple interest, how much after-tax interest will I earn? \\
- I am a recent graduate who has been employed for 6 months. If I save 300,000 KRW per month in a 2-year flexible savings product with a 7\% annual compound interest rate, how much will I accumulate for wedding expenses? \\
- I have about 400 million KRW in cash and plan to purchase an apartment by taking out a 400 million KRW mortgage. If the annual interest rate is 2.1\% with a 30-year equal principal-and-interest repayment plan, how much interest would I pay each month? \\ \\

\textbf{[Output Format]} Output the generated question in the following JSON format. \\
\texttt{\{} 
\texttt{"product": "<Product type (e.g., deposit, savings, loan)>", \textbackslash n }  
\texttt{"question": "<Generated question>"} 
\texttt{\}}
\\

\noalign{\hrule height 1pt}
\end{tabular}
}
\caption{Prompt for generating basic-level questions in BankMathBench.}
\label{tab:basic_prompt}
\end{table*}

\begin{table*}[t]
\centering
\scriptsize
\setlength{\tabcolsep}{4pt}
\renewcommand{\arraystretch}{1.12}
\resizebox{\textwidth}{!}{
\begin{tabular}{p{0.985\textwidth}}
\noalign{\hrule height 1pt}

\textbf{(Kor)}\\ 

\textbf{[역할]} 당신은 금융 전문가로서, 사용자가 제공한 조건을 바탕으로 예금, 적금, 대출 상품 중 2가지를 비교하는 금융 계산 질문을 생성하는 역할을 합니다. 아래 조건을 참고하여 2개의 금융 상품을 비교하는 질문을 *한 개만* 생성하세요. \\ \\

\textbf{[요구사항]} \\
- 질문은 상품별 금액, 이자율, 납입 방식, 기간 등을 모두 포함해야 하며, 최종적으로 ``어느 쪽이 이자가 더 높은가요?'', ``더 유리한 상품은 총 수령액이 얼마나 큰가요?'' 등 비교 목적이 명확히 드러나야 합니다. \\
- 비교 상품 유형: 예금, 적금, 대출 중 2가지 (예: 예금 vs 예금, 예금 vs 적금, 적금 vs 적금, 대출 vs 대출 등) \\
- 금액, 이자율, 기간 등의 조건은 매번 다르게 설정하고 동일한 조건이 반복되지 않도록 다양하게 생성하세요. \\
- 이자율: 1.0\%~9.9\% 범위에서 소수점 포함 랜덤 생성 \\
- 기간: 3개월~5년, 주 단위(26주, 52주 등) 또는 월 단위 포함 가능 \\
- 금액: 10만 원~1억 원 범위에서 다양하게 생성 \\
- 이자 방식: 단리 또는 복리 명시 \\
- 질문 형식: 비교 목적이 명확히 드러나는 표현 사용 \\ \\

\textbf{[예시]} \\
\# 상세 상품명: \\
(예금) 정기예금, 자유예금 \\
(적금) 26주적금, 자유적금 \\
(대출) 신용대출, 중신용대출, 사잇돌대출, 전월세보증금 대출, 마이너스통장 대출, 개인사업자 대출 \\ 
\\[-1em]
\# 질문 예시: \\
- 두 상품 간 총 이자액 차이가 궁금해요. \\
- 만기 수령액이 얼마나 차이 나나요? \\ \\

\textbf{[출력 형식]} 생성된 질문을 다음 JSON 형식으로 출력하세요. \\
\texttt{\{}  
    \texttt{"상품": "\textless 상품1 vs 상품2\textgreater", \textbackslash n }
    \texttt{"질문": "\textless 비교 질문\textgreater"} 
\texttt{\}} \\

\hline

\textbf{(Eng)}\\ 

\textbf{[Role]}
You are a financial expert responsible for generating a financial calculation question that compares two products among deposit, savings, and
loan products based on the conditions provided by the user.
Refer to the guidelines below and generate *only one* comparison question involving two financial products. \\ \\

\textbf{[Requirements]} \\
- The question must include the amount, interest rate, payment method, and term for each product, and clearly express the comparison objective,
such as “Which option yields higher interest?” or “How much greater is the total payout for the more advantageous product?" \\ 
- Product types to compare: choose any two among deposits, savings, and loans (e.g., deposit vs deposit, deposit vs savings, savings vs savings,
loan vs loan, etc.).  \\
- Vary the conditions (amount, interest rate, term) each time and avoid repeating identical configurations. \\
- Interest rate: randomly generate within 1.0\%–9.9\%, including decimals. \\
- Term: between 3 months and 5 years; weekly (e.g., 26 weeks, 52 weeks) or monthly units may also be used.  \\
- Amount: randomly generate between 100,000 KRW and 100,000,000 KRW. \\
- Interest calculation method: clearly specify simple or compound interest.\\
- The question must explicitly state the comparison objective. \\ \\

\textbf{[Example]} \\
\# Detailed Product Names: \\
(Deposit) Fixed deposit, Flexible deposit \\
(Savings) 26-week savings, Flexible savings \\
(Loan) Credit loan, Medium-credit loan, Saetdol loan, Jeonse deposit loan, Overdraft (minus account) loan, Business loan \\ 
\\[-1em]
\# Example Questions: \\
- I would like to know the total interest difference between the two products. \\
- How much difference is there in the maturity payout? \\ \\

\textbf{[Output Format]} Output the generated question in the following JSON format. \\
\texttt{\{} 
\texttt{"product": "\textless Product1 vs Product2\textgreater", \textbackslash n } 
\texttt{"question": "\textless Comparison question\textgreater"}
\texttt{\}}
\\
\noalign{\hrule height 1pt}
\end{tabular}
}
\caption{Prompt for generating intermediate-level questions in BankMathBench.}
\label{tab:intermediate_prompt}
\end{table*}

\begin{table*}[t]
\centering
\scriptsize
\setlength{\tabcolsep}{4pt}
\renewcommand{\arraystretch}{1.12}
\resizebox{\textwidth}{!}{
\begin{tabular}{p{0.985\textwidth}}
\noalign{\hrule height 1pt}
\textbf{(Kor)} \\
\textbf{[역할]}
다음 금융상품 조건을 바탕으로 금융 계산 문제를 *한 개만* 생성하시오. 일상생활에서 실제로 발생할 수 있는 현실적인 상황을 묘사하시오. \\ \\

\textbf{[조건]} \\
- 상품 유형: (예금 | 적금 | 대출) \\
- 납입 방식: (일시납 | 매월 | 원리금균등 | 만기일시상환 등) \\
- 이자 유형: (단리 | 복리 | 고정금리 등) \\
- 연이율: <Rate> \\
- 기간: (년 | 개월 | 주 | 일) \\
- 원금 또는 납입 금액: <금액> \\
- 세금: (이자소득세 15.4\% | 비과세) \\
- (선택)
  중도해지: <해지 가능 시점>, <중도 해지 (이율, 예정 시점)>  \texttt{\textbackslash n}
  중도상환: <중도 상환 (수수료율, 시점, 금액)>\\ \\

\textbf{[요구사항]} \\
- 불필요한 수치는 포함하지 마시오. \\
- 설명이나 해설, 정답은 포함하지 말고 문제만 출력하시오. \\
- 마지막 문장은 반드시 계산 가능한 금액을 명시적으로 질문하시오. \\
- 아래 목록에서 적절한 표현을 선택하여 사용하시오 (유사한 의미의 변형 표현 허용). \\ \\

\textbf{[예시]}\\
\# 표현 예시:\\
- (예금 | 적금) 총 원리합계는 얼마인가? \textbackslash n 세후 이자는 얼마인가? \textbackslash n 발생하는 이자 수익은 얼마인가? \ldots\ \texttt{(중략)}

- (대출) 총 상환 금액은 얼마인가? \textbackslash n 중도상환 수수료는 얼마인가? \textbackslash n 중도상환 후 남은 기간 동안 지급해야 할 이자는 얼마인가? \ldots\ \texttt{(중략)} \\
\\[-1em]
\# 용어 설명: \\
- (예금 | 적금) 총 이자, 이자 수익, 세후 이자: 이자만 \textbackslash n 총 수령액, 총 원리합계: 원금 + 이자 \\
- (대출) 총 이자, 이자 부담액: 이자만 \textbackslash n 총 상환액, 총 비용: 원금 + 이자 + 수수료 \\
\\[-1em]
\# 질문 예시: \\
- (예금) K씨는 연 3.2\% 단리 조건의 2년 만기 정기예금에 720만 원을 예치하였다. 이자는 만기 시 지급되며, 15.4\%의 세금이 공제된다. 6개월 이후부터는 2.1\%의 중도해지 이율이 적용된다. 만약 7개월 후 해지한다면 해당 시점의 총 원리합계는 얼마인가? \\
- (적금) B씨는 연 3.2\% 단리 조건의 2년 만기 적금에 매월 30만 원씩 납입하고 있다. 이자는 만기 시 지급된다. 1년 후 0.8\%의 중도해지 이율이 적용되고 15.4\%의 세금이 공제될 때, 그녀가 수령하게 될 총 금액은 얼마인가? \\
- (대출) C씨는 연 3.8\% 변동금리로 5년 만기 만기일시상환 방식의 1억 원을 대출받았다. 2년 후 금리는 남은 기간 동안 4.1\%로 인상되었다. 이자와 수수료를 포함한 총 비용은 얼마인가?  \ldots\ \texttt{(중략)} \\ \\

\textbf{[출력 형식]} 생성된 질문을 다음 JSON 형식으로 출력하세요. \\
\texttt{\{}  
    \texttt{"상품": "\textless 상품 유형 (예: 예금, 적금, 대출)\textgreater", \textbackslash n } 
    \texttt{"질문": "\textless 생성된 질문\textgreater"} 
\texttt{\}} \\
\hline
\textbf{(Eng)} \\
\textbf{[Role]}
Generate *only one* financial calculation question based on the following financial product conditions. Describe a realistic situation that could occur in everyday life. \\ \\

\textbf{[Conditions]} \\
- Product Type: (Deposit | Savings | Loan) \\
- Payment Method: (Lump-sum | Monthly | Amortized | Bullet repayment, etc.) \\
- Interest Type: (Simple | Compound | Fixed rate, etc.) \\
- Annual Interest Rate: <Rate> \\
- Term: (Years | Months | Weeks | Days) \\
- Principal or Installment Amount: <Amount> \\
- Tax: (15.4\% interest income tax | Tax-exempt) \\
- (Optional) Early Termination: <Eligible point>, <Early termination (rate, point)>, \texttt{\textbackslash n} Prepayment: <Prepayment (fee rate, point, amount)> \\ \\

\textbf{[Requirements]} \\

- Do not include unnecessary numerical values.\\
- Do not include explanations or answers. Output only the problem.\\
- The final sentence must explicitly ask for a calculable amount.\\
- Use an appropriate expression selected from the list below (variations with similar meaning are allowed).\\ \\

\textbf{[Example]}\\
\# Expressions:\\
- (Deposit | Savings) What is the total principal and interest? \textbackslash n How much interest will be earned after tax? \textbackslash n
 How much interest income will be generated? \ldots\ \texttt{(omitted)} \\
- (Loan) What is the total repayment amount? \textbackslash n What is the prepayment fee? \textbackslash n How much interest must be paid after prepayment for the remaining term? \ldots\ \texttt{(omitted)}\\ 
\\[-1em]
\# Terminology:\\
- (Deposit | Savings) Total interest, interest income, after-tax interest: interest only \textbackslash n Total payout, total principal and interest: principal + interest \\
- (Loan) Total interest, interest burden: interest only \textbackslash n Total repayment, total cost: principal + interest + fees \\ 
\\[-1em]
\# Example Questions:\\
- (Deposit) Mr. K deposited 7.2 million KRW into a two-year fixed deposit at an annual simple interest rate of 3.2\%. Interest is paid at maturity, and 15.4\% tax is deducted. Early termination is allowed after 6 months at a 2.1\% rate. If he terminates after 7 months, what is the total principal and interest at that time? \\
- (Savings)  Ms. B saves 300,000 KRW per month for two years at a 3.2\% annual simple interest rate. Interest is paid at maturity. If she terminates after one year at an early termination rate of 0.8\% with 15.4\% tax deducted, what is the total amount she receives? \\
- (Loan) Mr. C borrowed 100 million KRW at a 3.8\% annual variable rate for five years with a bullet repayment structure. After two years, the rate increased to 4.1\% for the remaining term. What is the total cost including interest and fees?  \ldots\ \texttt{(omitted)} \\ \\
 
\textbf{[Output Format]} Output the generated question in the following JSON format. \\
\texttt{\{} 
\texttt{"product": "<Product type (e.g., deposit, savings, loan)>", \textbackslash n } 
\texttt{"question": "<Generated question>"} 
\texttt{\}}
\\
\noalign{\hrule height 1pt}
\end{tabular}
}
\caption{Prompt for generating advanced-level questions in BankMathBench.}
\label{tab:advanced_prompt}
\end{table*}

\begin{table*}[t]
\centering
\scriptsize
\setlength{\tabcolsep}{4pt}
\renewcommand{\arraystretch}{1.12}
\resizebox{\textwidth}{!}{
\begin{tabular}{p{0.985\textwidth}}
\noalign{\hrule height 1pt}

\textbf{(Kor)}\\ 
\textbf{[역할]} \\  당신은 금융 계산을 정확히 수행하는 계산기입니다. \\
주어진 문제에 대해 다음 두 가지만 출력하세요: \\ \\

\textbf{[요구사항]} \\
1. 설명, 단위, 말풍선 등은 절대 포함하지 마세요. \\
2. 다양한 표현의 금액은 모두 계산 가능한 숫자값으로 표기하세요. \\
- 1백만2천 $\rightarrow$ 1002000 \\
- 30,000 $\rightarrow$ 30000 \\
- 오만 $\rightarrow$ 50000 \\ \\

\textbf{[참고 공식]} \\ 

(1) 예금 \\
- 단리 \\
이자 = 원금 $\times$ 연이율(\%) $\div$ 100 $\times$ (기간 $\div$ 365) \\ \\[-0.5em]

- 연 복리 기준 \\
만기금액 = 원금 $\times$ (1 + 연이율(\%) $\div$ 100)\texttt{\^{}}\,(기간(일) $\div$ 365) \\
이자 = 만기금액 - 원금 \\ \\[-0.5em]

(2) 적금 (정기적으로 동일 금액 납입) \\
- 단리 \\
이자 = 월납입액 $\times$ 납입개월수 $\times$ (납입개월수 + 1) $\div$ 2 $\times$ (연이율 $\div$ 12 $\div$ 100) \\ \\[-0.5em]

- 복리 \\
만기금액 = $\Sigma$ [월납입액 $\times$ (1 + 연이율 $\div$ 12 $\div$ 100)\texttt{\^{}}\,(납입개월수 - i)] \ (i = 0부터 n-1까지) \\
이자 = 만기금액 - 총납입액 \\ \\[-0.5em]

(3) 대출 \\
- 단리 \\
이자 = 대출원금 $\times$ 연이율 $\div$ 100 $\times$ (기간 $\div$ 365) \\ \\[-0.5em]

- 복리 (이자가 원금에 계속 합산될 경우) \\
상환금 = 대출원금 $\times$ (1 + 연이율 $\div$ 100)\texttt{\^{}}\,(기간(일) $\div$ 365) \\
이자 = 상환금 - 대출원금 \\ \\

\textbf{[출력 형식]} \\
\texttt{수식/Latex/코드: \textless계산에 사용한 수식/Latex/코드\textgreater} \\
\texttt{정답: \textless계산 결과 (숫자만)\textgreater} \\ \\

이제 아래 문제에 대해 위 형식으로 출력하세요: \textbackslash n \{\} \\

\hline
\textbf{(Eng)}\\
\textbf{[Role]} You are a calculator that performs financial calculations accurately. \\
For the given problem, output only the following two items: \\ \\

\textbf{[Requirements]} \\
1. Do not include explanations, units, speech bubbles, or any other text. \\
2. Convert all monetary expressions into numeric values that can be computed. \\
- ``1 million 2 thousand'' $\rightarrow$ 1002000 \\
- 30,000 $\rightarrow$ 30000 \\
- ``fifty thousand'' $\rightarrow$ 50000 \\ \\

\textbf{[Reference Formulas]} \\ 

(1) Deposits \\
- Simple interest \\
Interest = Principal $\times$ Annual rate(\%) $\div$ 100 $\times$ (Term $\div$ 365) \\ \\[-0.5em]

- Annual compounding \\
Maturity amount = Principal $\times$ (1 + Annual rate(\%) $\div$ 100)\texttt{\^{}}\,(Term(days) $\div$ 365) \\
Interest = Maturity amount - Principal \\ \\[-0.5em]

(2) Savings (equal periodic payments) \\
- Simple interest \\
Interest = Monthly payment $\times$ Number of months $\times$ (Number of months + 1) $\div$ 2 $\times$ (Annual rate $\div$ 12 $\div$ 100) \\ \\[-0.5em]

- Compound interest \\
Maturity amount = $\Sigma$ [Monthly payment $\times$ (1 + Annual rate $\div$ 12 $\div$ 100)\texttt{\^}\,(Number of months - i)] \ (i = 0 to n-1) \\
Interest = Maturity amount - Total payments \\ \\[-0.5em]

(3) Loans \\
- Simple interest \\
Interest = Loan principal $\times$ Annual rate $\div$ 100 $\times$ (Term $\div$ 365) \\ \\[-0.5em]

- Compound interest (when interest is continuously added to principal) \\
Repayment amount = Loan principal $\times$ (1 + Annual rate $\div$ 100)\texttt{\^{}}\,(Term(days) $\div$ 365) \\
Interest = Repayment amount - Loan principal \\ \\

\textbf{[Output Format]} \\
\texttt{Formula/LaTeX/Code: \textless formula/LaTeX/code used\textgreater} \\
\texttt{Answer: \textless computed result (numbers only)\textgreater} \\ \\

Now, for the problem below, output using the format above: \textbackslash n \{\} \\

\noalign{\hrule height 1pt}
\end{tabular}
}
\caption{Prompt for solution generation.}
\label{tab:solution_prompt}
\end{table*}

\begin{table*}[t]
\centering
\scriptsize
\setlength{\tabcolsep}{4pt}
\renewcommand{\arraystretch}{1.12}
\resizebox{\textwidth}{!}{
\begin{tabular}{p{0.985\textwidth}}
\noalign{\hrule height 1pt}

\textbf{(Kor)} \\
\textbf{[역할]} 당신은 주어진 질문, 수식/코드, 정답을 바탕으로, 한국어 줄글 형식의 리즈닝 과정을 작성하고 최종 정답을 출력하는 봇입니다. \\ \\

\textbf{[요구사항]}\\
1. 리즈닝 과정은 \texttt{<think>}와 \texttt{</think>} 태그 안에 작성합니다. \\
2. 리즈닝은 문제 풀이 과정을 서술형 한국어로 작성하며, 필요한 경우 간단한 수식을 포함합니다. \\
3. 수식/코드에 나타난 계산 로직을 분석하여, 같은 과정을 사람이 이해할 수 있는 풀이로 변환합니다. \\
4. 풀이는 핵심 내용만을 포함하여 간결하게 작성합니다. \\
5. 최종 정답은 \texttt{<answer>}와 \texttt{</answer>} 태그 안에 작성합니다. \\
6. 생성된 리즈닝 과정과 정답만 출력하세요. \\

\hline
\textbf{(Eng)}\\
\textbf{[Role]} You are a bot that generates a Korean narrative reasoning process and outputs the final answer based on the given question, equation/code, and answer. \\ \\

\textbf{[Requirements]} \\
1. Write the reasoning process inside the \texttt{<think>} and \texttt{</think>} tags. \\
2. The reasoning must describe the problem-solving process in narrative Korean, including simple formulas when necessary. \\
3. Analyze the computational logic shown in the equation/code and convert it into a human-readable solution that follows the same process. \\
4. Keep the explanation concise and focused only on the essential reasoning steps. \\
5. Write the final answer inside the \texttt{<answer>} and \texttt{</answer>} tags. \\
6. Output only the generated reasoning process and the final answer. \\

\noalign{\hrule height 1pt}
\end{tabular}
}
\caption{Prompt for generating reasoning and final answers.}
\label{tab:reasoning_prompt}
\end{table*}

\begin{table*}[t]
\renewcommand{\arraystretch}{1.3}
\centering
\resizebox{\textwidth}{!}{%
{\scriptsize 
\begin{tabular}{p{0.48\textwidth}|p{0.48\textwidth}}
\noalign{\hrule height 1pt}
\multicolumn{2}{l}{\textbf{Type 1: Formula Error}} \\
\hline
\textbf{(Kor)} \newline
\textbf{Question:} 2년 만기 정기예금에 연 3.5\% 복리로 500만원을 예금하는 것과,
매달 20만원씩 2년간 자유적금에 연 4.0\% 단리로 적금하는 것 중
만기 수령액이 얼마나 차이가 나나요?
&
\textbf{(Eng)} \newline \textbf{Question:} What is the difference in the maturity amount between depositing 5,000,000 KRW in a 2-year fixed-term deposit at an annual compound interest rate of 3.5\% and saving 200,000 KRW per month for 2 years in an installment savings account at an annual simple interest rate of 4.0\%?
\\ \\
\textbf{Zero-shot Output:} 
\newline
\texttt{<think> \ldots 다음으로, 자유적금의 만기 수령액을 계산해보겠습니다. 자유적금은 매달 20만원씩 2년간 예금하며, 연 4.0\% 단리로 이자를 받습니다.}

{\color{errorfg}\ttfamily
\textbf{[Error]} \newline \textbf{만기 수령액} \newline
= 월납입액*12*(1+연이율/100)\textasciicircum (납입연수)
\newline = 200000*12*(1+0.04)\textasciicircum 2
\newline
\textless calc\textgreater \textbf{200000*12*(1+0.04)\textasciicircum 2} \textless/calc\textgreater
\newline
\textless result\textgreater \textbf{4993920} \textless/result\textgreater
}

\texttt{\textcolor{goldfg}{
\textbf{[Gold]} \newline
총 이자 \newline = 월납입액*납입개월수*(납입개월수+1)/2*(연이율/12/100) \newline= 200000*24*(24+1)/2*(4/12/100) \newline= 200000
\newline 총 원금 = 200000 * 24
\newline 만기 수령액 = 200000*24+200000 = 5000000 \newline
\textless calc\textgreater \textbf{200000*24+200000*24*(24+1)/2*(4/12/100)} \textless/calc\textgreater
\newline
\textless result\textgreater \textbf{5000000} \textless/result\textgreater
}
}

\texttt{\textcolor{black}{\ldots </think> <answer> \ldots </answer>}}

& 
\textbf{Zero-shot Output:}
\newline
\texttt{<think> \ldots Next, we calculate the maturity amount of the installment savings account. The installment savings account involves depositing 200,000 KRW per month for 2 years and earns interest at an annual simple rate of 4.0\%.}

{\color{errorfg}\ttfamily
\textbf{[Error]} \newline \textbf{Maturity Amount} \newline
= Monthly Payment*12*(1+Annual Interest Rate/100)\textasciicircum (Number of Years)
\newline = 200000*12*(1+0.04)\textasciicircum 2
\newline
\textless calc\textgreater \textbf{200000*12*(1+0.04)\textasciicircum 2} \textless/calc\textgreater
\newline
\textless result\textgreater \textbf{4993920} \textless/result\textgreater
}

\texttt{\textcolor{goldfg}{
\textbf{[Gold]} \newline
Total Interest \newline = Monthly Payment*Number of Months*(Number of Months+1)/2*(Annual Interest Rate/12/100) \newline= 200000*24*(24+1)/2*(4/12/100) \newline= 200000
\newline Total Principal = 200000 * 24
\newline Maturity Amount = 200000*24+200000 = 5000000 \newline
\textless calc\textgreater \textbf{200000*24+200000*24*(24+1)/2*(4/12/100)} \textless/calc\textgreater
\newline
\textless result\textgreater \textbf{5000000} \textless/result\textgreater
}
}

\texttt{\textcolor{black}{\ldots </think> <answer> \ldots </answer>}}
\\

\noalign{\hrule height 1pt}

\multicolumn{2}{l}{\textbf{Type 2: Parameter Conversion Error}} \\
\hline
\textbf{(Kor)} \newline
\textbf{Question:} A씨는 2,229,900,000원을 일시불로 예치한 예금 상품을 가입했습니다. 기본 금리는 연 5.6\%이며, 이자는 단리로 계산됩니다. 이자소득세(15.4\%)가 공제됩니다. 13주 후 중도해지 가능하며, 중도해지 이율 3.36\%가 적용됩니다. 13주 후 중도해지 시 수령할 원금과 이자의 합계는 얼마인가요?
&
\textbf{(Eng)} \newline
\textbf{Question:} Mr. A deposited 2,229,900,000 KRW as a lump sum in a fixed deposit product. The base interest rate is 5.6\% per annum, and interest is calculated using simple interest. Interest income tax (15.4\%) is deducted. Early termination is allowed after 13 weeks, at which point an early termination interest rate of 3.36\% is applied. What is the total amount of principal and interest received if the deposit is terminated after 13 weeks?
\\ \\
\textbf{Zero-shot Output:}
\newline
\texttt{<think>1. 먼저 기본 금리로 13주 동안의 이자를 계산합니다.\textbackslash n
   2. 그 다음 이자소득세를 공제한 후 실제 수령 가능한 이자를 계산합니다.\textbackslash n
   3. 중도해지 이율을 적용하여 13주 후의 이자를 계산합니다.\textbackslash n
   4. 기본 이자와 중도해지 이자를 합하여 최종 수령액을 계산합니다.\textbackslash n
   5. 계산 결과를 정답으로 작성합니다. \ldots}\newline
\texttt{\textcolor{errorfg}{
\textbf{[Error]} \newline <calc>(2229900000*(\textbf{5.6}/100)*13/52)*(1-15.4/100)</calc>
}}

\texttt{\textcolor{goldfg}{
\textbf{[Gold]} \newline <calc> 2229900000+(2229900000*(\textbf{3.36}/100)*13/52)*(1-15.4/100) </calc>
}
}

\texttt{\textcolor{black}{\ldots </think> <answer> \ldots </answer>}}
&
\textbf{Zero-shot Output:}
\newline
\texttt{<think>1. First, calculate the interest for 13 weeks using the base interest rate.\textbackslash n
   2. Next, deduct the interest income tax to determine the actual receivable interest.\textbackslash n
   3. Apply the early termination interest rate to calculate the interest for 13 weeks.\textbackslash n
   4. Add the base interest and the early termination interest to compute the final amount received.\textbackslash n
   5. Present the calculated result as the final answer. \ldots}\newline
\texttt{\textcolor{errorfg}{
\textbf{[Error]} \newline <calc>(2229900000*(\textbf{5.6}/100)*13/52)*(1-15.4/100)</calc>
}}

\texttt{\textcolor{goldfg}{
\textbf{[Gold]} \newline <calc> 2229900000+(2229900000*(\textbf{3.36}/100)*13/52)*(1-15.4/100) </calc>
}
}

\texttt{\textcolor{black}{\ldots </think> <answer> \ldots </answer>}}
\\
\noalign{\hrule height 1pt}

\multicolumn{2}{l}{\textbf{Type 3: Arithmetic Error}} \\
\hline
\textbf{(Kor)} \newline
\textbf{Question:} A씨는 1,505,400,000원을 2년 만기 비과세 정기예금에 일시불로 예치했습니다. 기본 금리는 연 2.89\%이며, 이자는 복리로 계산됩니다. 만기 시점에 받게 되는 총 금액은 얼마인가요?
&
\textbf{(Eng)} \newline
\textbf{Question:} Mr. A deposited 1,505,400,000 KRW as a lump sum in a 2-year tax-exempt fixed-term deposit. The base interest rate is 2.89\% per annum, and interest is calculated on a compound basis. What is the total amount received at maturity?
\\ \\
\textbf{Zero-shot Output:}
\newline
\texttt{<think>A씨가 예치한 금액은 1,505,400,000원이고, 이자는 연 2.89\%이며 \ldots 
따라서, 총 금액은 \ldots <calc>1,505,400,000 * (1.0289)\textasciicircum 2 </calc>}

\texttt{\textcolor{errorfg}{
\textbf{[Error]} \newline <result>
\textbf{1,581,999,996.18}
</result>
}}

\texttt{\textcolor{goldfg}{
\textbf{[Gold]}  \newline <result>
\textbf{1,593,669,445.134}
</result>
}
}

\texttt{\textcolor{black}{\ldots </think> <answer> \ldots </answer>}}
&
\textbf{Zero-shot Output:}
\newline
\texttt{<think>The deposited amount is 1,505,400,000 KRW, and the annual interest rate is 2.89\% \ldots 
Therefore, the total amount is \ldots <calc>1,505,400,000 * (1.0289)\textasciicircum 2 </calc>}

\texttt{\textcolor{errorfg}{
\textbf{[Error]}  \newline <result>
\textbf{1,581,999,996.18}
</result>
}}

\texttt{\textcolor{goldfg}{
\textbf{[Gold]}  \newline <result>
\textbf{1,593,669,445.134}
</result>
}
}

\texttt{\textcolor{black}{\ldots </think> <answer> \ldots </answer>}}
\\
\noalign{\hrule height 1pt}
\end{tabular}%
} 
}
\caption{Illustrative Examples of Error Types}
\label{tab:error_type}
\end{table*}

\end{document}